\pdfoutput=1

\documentclass[11pt]{article}

\usepackage{acl}

\usepackage{amsmath,amsfonts,bm}

\def\eqref#1{equation~\ref{#1}}

\def\1{\bm{1}}

\DeclareMathAlphabet{\mathsfit}{\encodingdefault}{\sfdefault}{m}{sl}
\SetMathAlphabet{\mathsfit}{bold}{\encodingdefault}{\sfdefault}{bx}{n}

\usepackage{times}
\usepackage{latexsym}
\usepackage{diagbox}
\usepackage{multirow}
\usepackage{multicol}
\usepackage{hhline}
\usepackage{bbm}
\usepackage[T1]{fontenc}

\usepackage[utf8]{inputenc}

\usepackage{microtype}

\usepackage{inconsolata}
\usepackage{color}
\usepackage{url}
\usepackage{natbib}
\usepackage{caption}
\usepackage{subcaption}
\usepackage{csquotes}
\usepackage{xcolor,soul}
\usepackage{graphicx}
\usepackage{wrapfig}
\usepackage{comment}
\usepackage{tabularray}
\usepackage{algorithm}
\usepackage{rotating}

\newcommand\blfootnote[1]{%
  \begingroup
  \renewcommand\thefootnote{}\footnote{#1}%
  \addtocounter{footnote}{-1}%
  \endgroup
}

\title{BloomVQA: Assessing Hierarchical Multi-modal Comprehension}

\author{Yunye Gong$^{1,*}$, Robik Shrestha$^{2,*}$, Jared Claypoole$^1$, \\\textbf{Michael Cogswell$^1$, Arijit Ray$^{3,\dagger}$, Christopher Kanan$^{4,\ddagger}$ and Ajay Divakaran$^1$}\\$^1$SRI International\\$^2$Department of Computer Science, Rochester Institute of Technology\\ $^3$Department of Computer Science, Boston University\\
$^4$Department of Computer Science, University of Rochester\\
\tt\small$^1${first.last@sri.com},$^2${rss9369@rit.edu},$^3${array@bu.edu},$^4${ckanan@cs.rochester.edu}}

\begin{document}
\maketitle
\blfootnote{$^{*}$: These authors contributed equally to this work}
\blfootnote{$^{\dagger}$: work completed at SRI International}
\blfootnote{$^{\ddagger}$: work completed at Rochester Institute of Technology}
\begin{abstract}
We propose a novel VQA dataset, BloomVQA, to facilitate comprehensive evaluation of large vision-language models on comprehension tasks. Unlike current benchmarks that often focus on fact-based memorization and simple reasoning tasks without theoretical grounding, we collect multiple-choice samples based on picture stories that reflect different levels of comprehension, as laid out in Bloom's Taxonomy, a classic framework for learning assessment widely adopted in education research. Our data maps to a novel hierarchical graph representation which enables automatic data augmentation and novel measures characterizing model consistency. We perform graded evaluation and reliability analysis on recent multi-modal models. In comparison to low-level tasks, we observe decreased performance on tasks requiring advanced comprehension and cognitive skills with up to 38.0\% drop in VQA accuracy. In comparison to earlier models, GPT-4V demonstrates improved accuracy over all comprehension levels and shows a tendency of bypassing visual inputs especially for higher-level tasks. Current models also show consistency patterns misaligned with human comprehension in various scenarios, demonstrating the need for improvement based on theoretically-grounded criteria. The dataset can be accessed at \url{https://huggingface.co/datasets/ygong/BloomVQA}.
\end{abstract}
\section{Introduction}
Recent advances of machine intelligence solutions have demonstrated tremendous success in a wide range of language and multi-modal tasks over diverse domains~\cite{foundation,gpt3,gpt4,chatgpt,llama2}. With increasing popularity of such solutions, how to systematically evaluate and analyze the models remains a key concern for confident application. Many recent efforts have focused on probing the capabilities and risks of the models based on different tasks, measures and perspectives~\cite{evaluate_survey,bm1,bm2,bm3}. For example, recent studies have explored functionalities and limitations of large language models (LLMs) from the perspectives of cognitive science~\cite{cognitive} and semantic 
consistency~\cite{sahu22,vconcept}.\par
\begin{figure}[t]
\hspace{-0.25em}
\includegraphics[scale=0.38]{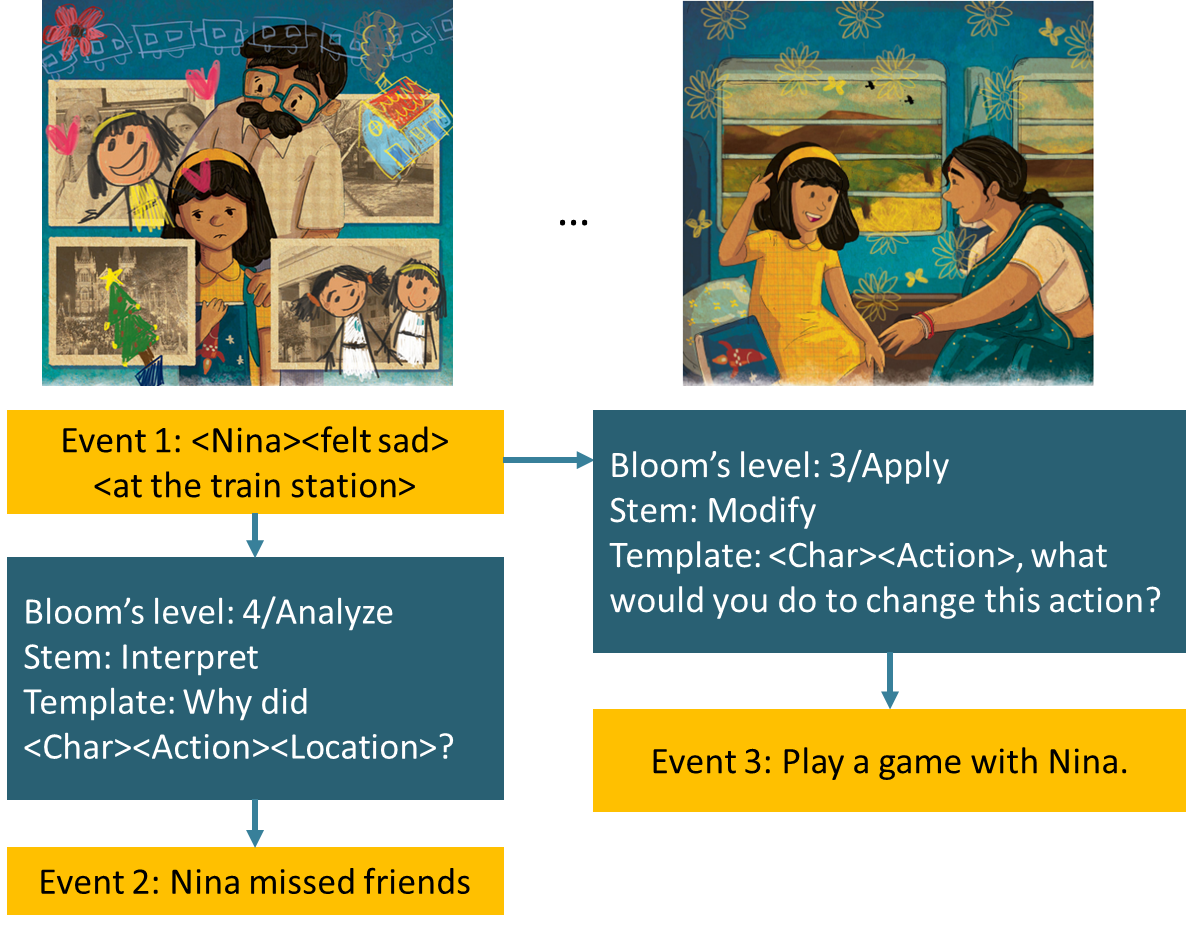}
\caption{Story graph: a hierarchical graph representation based on Bloom's Taxonomy} 
\label{fig:teaser}
\vspace{-1.5em}
\end{figure}
\begin{figure*}[t]
\hspace{-0.5em}
\centering
\begin{tabular}{cc}
\includegraphics[scale=0.3]{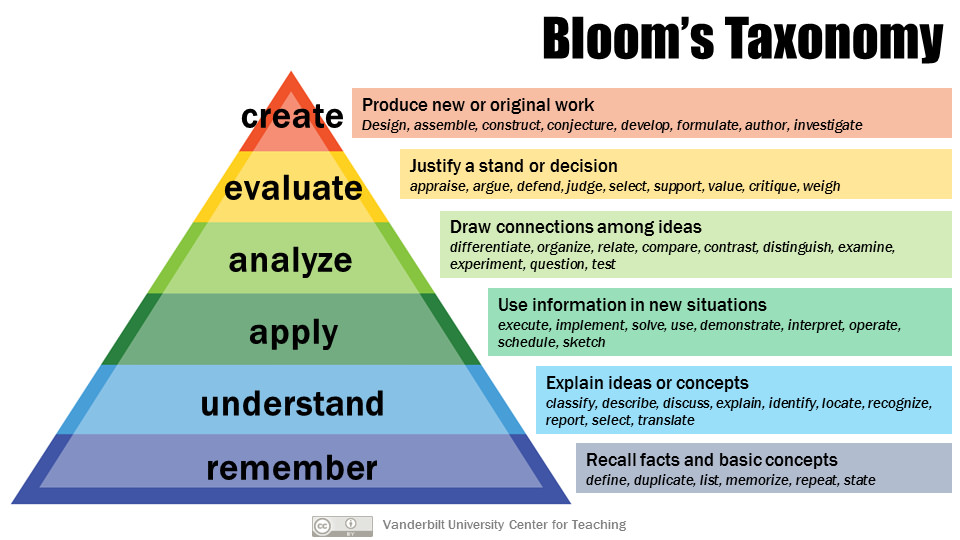}
& \includegraphics[scale=0.45]{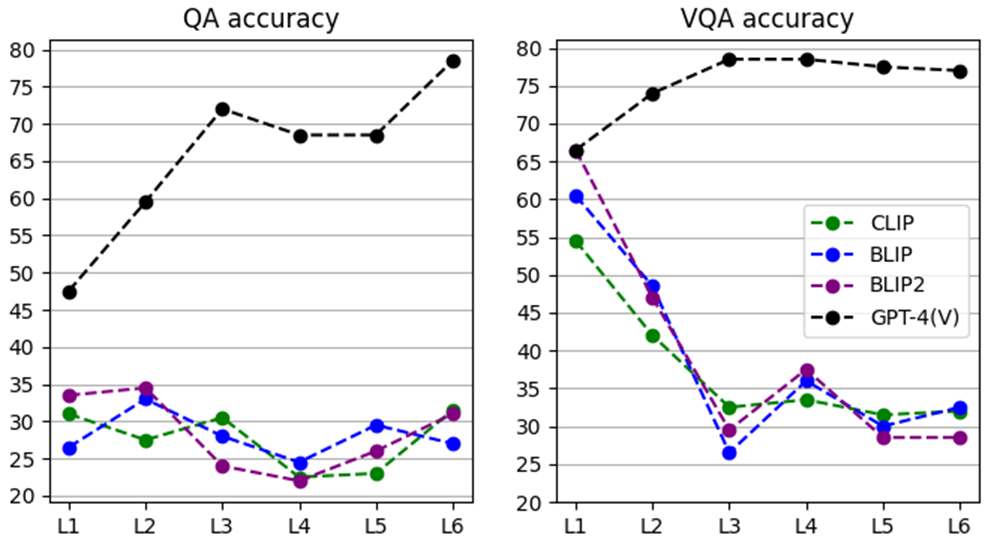}
\end{tabular}
\vspace{-1em}
\caption{Graded evaluation on BloomVQA data following Bloom's Taxonomy~\cite{bloom_figure}. For VLP models, the VQA accuracy decreases as the task level increases, while the QA accuracy using no visual inputs remains low. For GPT-4V, the VQA accuracy greatly improves over all levels while the comparison to QA accuracy suggests that the model tends to either bypass or even get confused by visual contents especially at high levels.} 
\label{fig:bloom}
\vspace{-1em}
\end{figure*}
In the context of Visual Question Answering (VQA), several benchmarks target demonstration of multi-modal reasoning and comprehension beyond mere memorization of low-level statistical patterns in the data. For example, GQA~\cite{GQA} leverages scene graph representations to construct VQA tasks describing reasoning based on relations of different entities in single-frame images. Other datasets~\cite{movieqa,recipeQA} leverage sequential visual inputs to form evaluations addressing procedural tasks or story comprehension over longer time horizons. However, little effort has been made in curating datasets to support systematic evaluation of multi-modal comprehension following a theoretically-grounded definition of comprehension, leading to tasks describing only a limited set of cognitive processes at relatively low levels. Motivated by this challenge, we draw inspiration from theories and methodologies developed by education researchers on K-12 reading comprehension, where comprehension tasks are graded based on underlying cognitive processes involved following a taxonomy named Bloom's Taxonomy~\cite{bloom1956,bloom2001}.\par
In Bloom's Taxonomy (Figure~\ref{fig:bloom}), handling of knowledge is categorized into 6 levels from the most basic (Level 1) to the most advanced (Level 6), including ``remember,'' ``understand,'' ``apply,'' ``analyze,'' ``evaluate'' and ``create.'' For each level, a group of cognitive skills are identified as required to handle tasks at the corresponding level. For example, the skill of differentiating is required to achieve successful analysis (Level 4), while the skill of critiquing is required to achieve successful evaluation (Level 5). Following this taxonomy, we propose BloomVQA dataset containing VQA tasks categorized based on the level of cognitive skills required. To support a wide range of tasks reflecting basic and advanced comprehension, we collect multiple-choice VQA samples with open-ended questions and free-form answer choices based on picture stories designed for early childhood education. Serving as an evaluation set for multi-modal comprehension, BloomVQA contains 1200 core data samples collected from human annotators based on 20 stories, categorized into 6 Bloom's levels. The core data samples are further augmented based on the hierarchical structure of Bloom's Taxonomy. Specifically, we generalize the concept of scene graphs and propose a novel graph representation named the Story Graph (Figure~\ref{fig:teaser}). While scene graphs focus on attributes and relations of low-level entities mapped to knowledge mostly at Level 1 in Bloom's Taxonomy, to accommodate advanced tasks at higher Bloom's levels, we consider different events, each of which can presumably be described by a separate scene graph, as the nodes of a Story Graph. We consider edges describing relations between pairs of events. In addition to low-level relations (e.g., temporal relations), we describe relations corresponding to underlying cognitive processes (e.g., making inference, making prediction) which link different events in reaching an overall comprehension of the story. In this way, given a VQA task constructed about one event, new VQA tasks can be automatically constructed at a combinatorial scale by traversing the underlying Story Graph, as information recorded along the path of traversal can be introduced as the context in forming new VQA tasks.\par
BloomVQA can be used as a benchmark for multi-modal comprehension engines including current multi-modal LLMs~\cite{gpt4,llava,minigpt4}, as it enables systematic evaluation of model capabilities with quantitative metrics characterizing its alignment to human comprehension, which is an important aspect for models to approach Artificial General Intelligence (AGI)~\cite{agi}. First of all, with data categorized based on Bloom's levels, we can probe models in a graded manner and examine model behaviors on tasks posing different levels of challenges for human comprehension, especially including tasks requiring high-level cognitive skills. Furthermore, we propose novel metrics characterizing the consistency of model performance following the underlying hierarchy and examine intuitive hypotheses mapping model behaviors to human comprehension patterns. In this work, we examine the hypothesis that models succeed at more challenging tasks are less likely to fail at easy ones. We also study how relevant knowledge, where the relevancy can be defined based on the underlying taxonomy, would affect model performance. For example, lower-level knowledge which can be comprehended using basic cognitive skills (e.g., memorizing "what is the name of the girl?") is unlikely to have a big effect on solving tasks that require more advanced cognitive skills (e.g., making inference about "what would happen if the girl is taking a flight rather than a train?"). In addition to being accurate, being consistent and consequently predictable is critical for models to be applied by human users with confidence.\par
Using the proposed dataset and metrics, we conduct comprehensive evaluation of recent Vision-Language Pre-training (VLP) models~\cite{CLIP,BLIP,BLIP2} and a state-of-the-art multi-modal LLM~\cite{gpt4}. 
We observe that for all three VLP models, the comprehension accuracy degrades for tasks requiring higher-level cognitive skills, in comparison to tasks requiring mere memorization of low-level details that can be directly identified from either textual or visual data. With zero-shot prompting experiments using GPT-4V~\cite{gpt4}, we observe generally higher VQA accuracy over all levels. However, there is also an increasing tendency for the model to take shortcuts and bypass visual comprehension when considering tasks requiring higher-level comprehension, as shown in Figure~\ref{fig:bloom}. Furthermore, via consistency analysis, we observe that model behaviors deviate from human comprehension patterns and intuitions at various scenarios. The results demonstrate the need for examining and improving current solutions based on theoretically-grounded criteria of reasoning and comprehension.
In summary, our core contributions include the following:
\begin{itemize}
\item
We present a novel BloomVQA dataset for systematic evaluation of multi-modal comprehension. The dataset is constructed based on Bloom's Taxonomy, a classic framework from education research which provides categorization of learning and comprehension. 
\item
We propose the Story Graph, a novel hierarchical graph representation based on Bloom's Taxonomy for story comprehension which enables meaningful data augmentation via traversing the graph.
\item
We propose novel metrics based on the underlying data hierarchy to examine the consistency of model performance for multi-faceted analysis on model reliability.
\item 
We evaluate state-of-the-art models with our proposed dataset and metrics. We show that improvement of current solutions is needed especially to address tasks reflecting higher-level cognitive skills and to demonstrate consistent and reliable comprehension.
\end{itemize}
\section{Related Works}

\paragraph{Bloom's Taxonomy}
First established by~\citet{bloom1956} and further revised by~\citet{bloom2001}, Bloom's Taxonomy describes a framework categorizing learning objectives based on levels of complexity of underlying cognitive processes. There are six major categories, each represented by a set of skills and actions describing the corresponding cognitive processes.
Specifically, direct recall of concrete facts and knowledge is considered as the basis (Level 1) of objectives requiring higher-order reasoning and abstraction, from understanding and being able to explain (Level 2) to creating and generating original works (Level 6). While Bloom's Taxonomy has been widely adopted in K-12 and college education as the guidance for designing teaching instructions and assessments~\cite{Thompson08,Shuhidan09}, limited effort has been made to transfer the taxonomy to the machine learning domain as the guidance for analyzing, assessing and improving learning. 
\citet{zhang21} propose a learning-based model for automatic classification of education questions based on Bloom's Taxonomy. 
\citet{sahu21} use Bloom's Taxonomy to form proximal clarifying contexts and study their impact on language models. In this work, we propose a novel VQA dataset to support hierarchical model assessment guided by Bloom's Taxonomy. Specifically, we use the action verbs associated with different Bloom's levels to create templates for data collection. While it is noted in recent studies that using the action verbs alone may not be sufficient in fully describing the underlying learning process~\cite{bloom_flaw}, the verbs provide us a concrete operational definition of the taxonomy as the guidance for understanding a coarse trend of model performance over data categorization.
\paragraph{Visual Comprehension and Reasoning}
Motivated by the multi-modal nature of human comprehension and communication~\cite{multimodal}, many recent datasets seek to challenge VQA models by tasks requiring reasoning in different domains~\cite{data_survey, garg22, chris1,chris2}. Several works introduce tasks requiring understanding based on multiple images. MovieQA~\cite{movieqa} dataset contains large scale multiple-choice samples collected for story comprehension based on video clips. RecipeQA~\cite{recipeQA} is constructed based on recipes with procedural text and image instructions to evaluate visual understanding over events with temporal relations in the form of clozing and ordering tasks. SlideVQA~\cite{slidevqa} proposes data and tasks requiring multi-hop reasoning over slide decks with multiple pages. GQA~\cite{GQA} leverages Visual Genome annotations~\cite{genome} to introduce tasks requiring reasoning with respect to entity relations in scene graphs.\par 
In comparison to existing data where categorization is either based on the prefix of the question or intuitive summary of the tasks, we provide principled and theoretically-grounded categorization based on Bloom's Taxonomy. Furthermore, we generalize the concept of scene graphs and propose the Story Graph which integrates entity-level relations, temporal relations and relations defined by the cognitive processes from Bloom's Taxonomy, enabling hierarchical data augmentation and comprehensive model evaluation with novel metrics.  
\section{BloomVQA Dataset}
\paragraph{Data Collection}
We propose a novel dataset for systematic assessment of multi-modal comprehension. We collect picture stories designed for educating young children from two Creative Commons~\cite{cc4} resources~\cite{storyweaver,bookdash} which provide collections emphasizing an Indian and an African cultural background respectively. 
We manually select 20 stories with relatively consistent artistic style, length (around 10-20 pages) and plot complexity so that tasks reflecting different levels of human comprehension and requiring different types of cognitive skills can be meaningfully collected based on the content of the stories. We filter out the stories with distorted artistic styles as they can pose additional challenges to both human annotators and learning models.\par
We provide a web-based UI (details included in the Appendix) where an annotator is asked to read through a picture story and then provide a set of 6 multiple-choice samples in English, each corresponding to one level in Bloom's Taxonomy. In Bloom's Taxonomy, each level of comprehension is associated with a set of action verbs describing underlying cognitive processes and skills (e.g., "identify" for Level 1). 
We use the action verbs to construct a set of more than 70 question templates as our operational definition of Bloom's Taxonomy. For example, "How would you compare the <attribute> of <character> at the beginning and the end of the story?" is provided as a template based on the Level 4 action verb "compare". More examples are included in the Appendix.
For each sample, we collect one template-based question and 4 free-form answers including 1 correct answer and 3 incorrect answers from the same annotator.\par
We collect data samples via Amazon Mechanical Turk (AMT) and manually review all inputs to select data with proper reflection of the story and the underlying Bloom's level. For each story, we collect 10 sets of reviewed inputs from different annotators. To further reduce the bias in the dataset, we enforce annotators to provide answer choices with consistent lengths for the same question, with a length difference fewer than 5 words. 

\begin{table}[H]
\centering
\small
\setlength\tabcolsep{5pt}
\begin{tabular}{c|c|c|c}
\hline
Level & \shortstack{\\Question} & \shortstack{\\Correct} & \shortstack{\\Incorrect}  \\
\hline
\shortstack{\\1} & 10.83$\pm$ 2.24 & 2.84$\pm$ 1.80 & 2.71$\pm$ 1.67 \\
\shortstack{\\2} & 13.74$\pm$ 2.86 & 5.23$\pm$ 2.35 & 5.04$\pm$ 2.28\\
\shortstack{\\3} & 14.08$\pm$ 2.60 & 5.39$\pm$ 2.33 & 5.16$\pm$ 2.26\\
\shortstack{\\4} & 11.69$\pm$ 3.97 & 6.72$\pm$ 3.57 & 6.59$\pm$ 3.59\\
\shortstack{\\5} & 11.43$\pm$ 3.42 & 6.16$\pm$ 2.98 & 6.11$\pm$ 3.04\\
\shortstack{\\6} & 16.68$\pm$ 3.05 & 6.23$\pm$ 3.02 & 5.86$\pm$ 2.80\\
\hline
\end{tabular}
\caption{Data statistics: average and standard deviation of data length.
}
\label{tab:length}
\end{table}
Detailed statistics about the data are shown in Table~\ref{tab:length}. In contrast to VQA data considering simple answers, we collect data with long free-form answers especially for high-level comprehension tasks. Overall we collect a core set of 1200 multiple-choice samples, based on which we perform systematic data augmentation as described in Sec~\ref{sec:aug}. We demonstrate consistency analysis with a set of 12k augmented samples.

We perform human baseline on proposed dataset (1200 core VQA samples) with a small group of adult reviewers. The human baseline reports an average accuracy of $89\%$ with $2\%$ standard deviation. This preliminary study provides insights on the current gap between human and machine capabilities (GPT-4V reports an average accuracy of $75.3\%$ as shown in Table~\ref{tab:base}) on the proposed dataset.

\paragraph{Story Graph}\label{sec:storygraph}
Grounding VQA data to an established taxonomy provides not only theory-based categorization, but also guidance on systematic probing and generation of the data. Based on Bloom's Taxonomy, we propose a hierarchical graph representation for picture story data. Referred to as the Story Graph, this novel representation naturally extends the concept of scene graphs~\cite{genome}. While scene graphs focus on representing low-level information such as geometric relations between local entities based on individual images, in a Story Graph (Figure~\ref{fig:teaser}),  we represent different events in a single story as the nodes and represent event-level relations as the edges. In addition to low-level relations (e.g., temporal relations), we include a wide range of higher-level relations (e.g., logical relations, causal relations) corresponding to different levels of cognitive skills specified in Bloom's Taxonomy (e.g., making comparison, making inference). As shown in Fig.~\ref{fig:teaser}, each of our templates correspond to one type of the edges in the Story Graph describing a particular cognitive skill. The type of edge is the same for all instantiations of the template. For example, the template "<Character><Action>, what would you do to change this action? <Answer>" always creates a Level 3 edge between two events, each encoding Level 1 details such as the character involved and the action taken. In this way, rich and diverse knowledge about the story requiring different levels of comprehension is organized in a hierarchical manner. 
The benefits of the proposed Story Graph are multi-fold. By traversing through the underlying graph, we can achieve systematic augmentation of VQA data following the taxonomy encoded in the graph. For example, given a task about a base event, we can construct an augmented question by incorporating knowledge about each connected event as the context. This augmentation not only expands the dataset combinatorially, but also introduces novel metrics for assessing the consistency of learning models. As context information introduced in the augmentation is labeled with Bloom's levels, consistency analysis on how context knowledge requiring different levels of comprehension would affect base tasks can be performed to characterize model reliability.

\section{Experimental Analysis}
To demonstrate the multi-faceted assessment enabled by our framework, we perform zero-shot experiments using three recent VLP models including CLIP~\cite{CLIP}, BLIP~\cite{BLIP} and BLIP2~\cite{BLIP2} with reported implementations~\cite{lavis}. We further perform zero-shot prompting-based experiments using the state-of-the-art GPT-4V~\cite{gpt4}. The baseline models are proposed to demonstrate a wide range of vision-language tasks probing multi-modal comprehension. 
For CLIP experiments, we use pretrained ViT-B-32 image transformer. For BLIP and BLIP2 experiments, we use pretrained models in the image-text matching (ITM) mode~\cite{BLIP2}. The models contain 583M and 188M trainable parameters as reported in the original papers. We use one Nvidia Tesla V100 GPU for experiments with VLP models. 
\paragraph{Multiple-choice Accuracy}
We first evaluate the models with our core dataset of 1200 multiple-choice samples under two settings. In the first setting, we perform the "Hasty Student" experiment~\cite{movieqa} by predicting multiple-choice answers solely based on the question text. This baseline measures underlying biases, as high accuracy indicates that the model is making up answers effectively without reading the actual picture stories. 
For experiments with three VLP models, we extract text embeddings of the question $q$ and each answer candidate $a_i$ from the set of answer choices $\{a_i\}_{i=1}^{N=4}$ using the text encoder and examine cosine similarity between the question embedding and each candidate answer embedding. We select the answer corresponding to the largest QA similarity as our prediction.\par
In the second setting, we perform the "Searching Student" experiment~\cite{movieqa} by predicting the multiple-choice answer based on the cross-modal similarity defined as
\begin{align}
l(\{v_j\}_{j=1}^K,q,a_i) = \max_j(g(q,v_j)+g(a_i,v_j))\nonumber
\end{align}
where $g$ denotes the cosine similarity between normalized embeddings extracted from the text input (question $q$ and answer candidate $a_i$) and the image frames in the picture story where $\{v_j\}_{j=1}^K$ denotes a set of image frames from a story with $K$ pages and $v_j$ denotes the $j^{th}$ frame. We adopt max pooling for aggregating the similarity scores over image frames based on empirical performance of baselines. Other aggregation methods such as average pooling are tested with only minor performance differences noted. 
We select $a_{i^\star}$ with $i^\star=\arg\max_i l(\{v_j\}_{j=1}^K,q,a_i)$ as the model prediction.\par
\paragraph{Model Consistency: Conditional Performance} 
Categorizing tasks based on Bloom's Taxonomy enables evaluation of the consistency of model performance over tasks at different levels. As for humans, being consistent on relevant tasks is an important sign of comprehension on given subjects. In this work, we examine an intuitive hypothesis: models which succeed at more challenging tasks demonstrating comprehensive cognitive skills should be less likely to fail at easier ones requiring only basic skills. Let $X_m$ and $X_n$ denote a pair of VQA tasks from the same story and the same annotator at Bloom's levels $m$ and $n$ respectively. We compute the likelihood of having a model solving $X_m$ correctly given that the model is correct on $X_n$ as
\begin{align}
    P_{m,n} = \frac{1}{|S|}\sum_{s=1}^{|S|}\mathbbm{1}\{\hat{a}_{m,s}=\bar{a}_{m,s}|\hat{a}_{n,s}=\bar{a}_{n,s}\}\nonumber,
\end{align} where $|S|$ denote the total number of annotation sets. $\hat{a}$ and $\bar{a}$ denote the predicted answer and correct answer for a task respectively. If $P_{m,n}$ with $m<n$ is higher than the unconditional accuracy at level $m$, it suggests that the model performance complies with the intuitive hypothesis. In other words, for a model demonstrating consistent comprehension patterns resembling human behavior, its probability of solving easier tasks from a lower Bloom's level $m$ is expected to be high when the model is correct on more comprehensive tasks from a higher Bloom's level $n$. Note that the formulation of consistency as conditional performance can be further expanded based on different hypotheses to examine various forward and backward consistency patterns~\cite{yangyi}.
\paragraph{Model Consistency: Augmentation}\label{sec:aug}
Based on the data augmentation strategy described in Sec.~\ref{sec:storygraph}, we propose another strategy for probing model consistency by comparing its performance on visual comprehension tasks with and without background knowledge introduced. We consider the relevancy of background knowledge to a given task to be associated with the node connection in an underlying Story Graph constructed following Bloom's Taxonomy. For a consistent model, its likelihood of success on a task should only be improved when relevant background information is provided and should not be affected by irrelevant information. Correspondingly, limited information from a lower-level task is less likely to affect the model performance on a higher-level task requiring comprehensive understanding of the story. While background information from a higher-level task involving advanced cognitive skills and reasoning is more likely to be useful for inferring the answer of a lower-level task.
For example, considering a Level 3 base task ("If you are Nina's father, what would you do to change her mood?") which requires distilling information from the story for problem solving, background information from a Level 1 context task ("What's Nina's father's hair color? Black.") is less likely to be helpful. Therefore, a consistent model would have a similar performance on the augmented task ("If you are Nina's father, who has black hair, what would you do to change her mood?") given the same set of candidate answers. Meanwhile, the background information from a Level 4 context task ("Why is Nina sad? Because she misses friends.") is expected to facilitate comprehension, leading to improved performance on the augmented task ("Nina is sad because she misses friends. If you are Nina's father, what would you do to change her mood?").\par
 Given a context task $X_n$ and a base task $X_m$ drawn from the core dataset at level $n$ and $m$ respectively, we construct an augmented task $X_{m|n}$ by prepending the question $q_n$ and the correct answer $\bar{a}_n$ of the context task to the base question $q_m$. We keep answer choices of the base task $a_m$. For experiments with three VLP models, we extend the Searching Student formulation for the augmented task by defining the cross-modal matching score between each candidate answer choice $a_{m,i}$ and the picture story with frames $\{v_j\}_{j=1}^K$ as 
\begin{align}
 l(\{v_j\}_{j=1}^K,q_n,\bar{a}_n,q_m,a_{m,i}) & = max_j(g(q_n,v_j)\nonumber\\
 +g(\bar{a}_n,v_j) +g(q_m,v_j)& + g(a_{m,i},v_j))\nonumber.
\end{align} 
We select the answer corresponding to the highest score as the prediction for the augmented task. 
While combinatorial augmentation can be achieved, in this work, we examine a set of 12k augmented samples considering Level 1 data of the same story as the context task. Let $Y_{base}$ denote the model performance as a binary score for each base task without augmentation. Let $Y_{aug}$ denote the model performance as the average accuracy on the set of augmented tasks constructed by incorporating low-level contexts. We compute consistency defined as the average precision between two sets of scores~\cite{sahu22}: $AP(Y_{base},Y_{aug})$. This metric quantifies the predictability of model performance when probing the knowledge along the underlying hierarchical graph. 

\paragraph{Experiments using GPT-4V}
We perform zero-shot prompting experiments using the \verb|gpt-4-1106-preview| model and the \verb|gpt-4-vision-preview| model. In comparison to experiments using VLP models, we ask GPT-4V model to make a prediction based on all images in each story at once. We perform experiments on the core dataset with 1200 samples and a set of 1200 augmented data constructed by considering a random Level 1 context task from the same story for each base sample from the core dataset. 
We perform computation via OpenAI API. The total computation corresponds to approximately 8M tokens. 
We use the following prompts for multiple-choice evaluation. For each task, we randomly order the candidate answers and parse the output of LLM to compare the index generated following ``My chosen answer is'' to the ground-truth index of the correct answer.
For GPT-4 experiments, we use the prompt:
\begin{quote}
Choose the best answer based on the question. End your response with `My chosen answer is' followed by your chosen answer.\\
<Question>\\
<Candidate Answers>
\end{quote}
For GPT-4V experiments, we use the prompt:
\begin{quote}
<Images>\\
Choose the best answer based on the story in the images. End your response with `My chosen answer is' followed by your chosen answer.\\
<Question>\\
<Candidate Answers>
\end{quote}

With prompts specified above, we receive indeterminate answers for 5\% and 2\% of the samples in QA-only and VQA experiments respectively. We report the average performance on the rest of the samples in Table~\ref{tab:base}-\ref{tab:aug}.
\begin{table}[H]
\centering
\small
\vspace{-1em}
\setlength\tabcolsep{3pt}
\centering
\begin{tabular}{c c c|c c|c c|c c}
\hline
 &\multicolumn{2}{c|}{\shortstack{\\CLIP}}
&\multicolumn{2}{c|}{\shortstack{\\BLIP}}&\multicolumn{2}{c|}{\shortstack{\\BLIP2}}&\multicolumn{2}{c}{\shortstack{\\GPT-4V}}\\\hline
Level &\shortstack{\\QA}&\shortstack{\\VQA}&\shortstack{\\QA}&\shortstack{\\VQA}&\shortstack{\\QA}&\shortstack{\\VQA}&\shortstack{\\QA}&\shortstack{\\VQA}\\
\hline
\shortstack{\\1} & 31.0 & 54.5 & 26.5 & 60.5 & 33.5 & 66.5 &47.5 &66.5 \\
2 & 27.5 & 42.0 &  33.0 & 48.5 &34.5 & 47.0 &59.5 &74.0\\
3 & 30.5 & 32.5 & 28.0 & 26.5 &24.0 & 29.5 & 72.0 & 78.5\\
4 & 22.5 & 33.5 & 24.5 & 36.0 & 22.0 & 37.5& 68.5 & 78.5\\
5 & 23.0 & 31.5 &29.5 & 30.0 &26.0 & 28.5 & 68.5 &77.5\\
6 & 31.5 & 32.0 & 27.0 & 32.5 & 31.0 & 28.5 & 78.5 & 77.0\\
\hline
\shortstack{\\Avg.} & 27.7 & 37.6 & 28.0 & 39.0 & 28.5 & 39.6 & 65.8 & 75.3\\
\hline
\end{tabular}
\caption{Accuracy (\%) on 1200 BloomVQA samples.}\label{tab:base}
\end{table}
\section{Results and Discussion}
\paragraph{Multiple-choice Accuracy}
In Table~\ref{tab:base}, we present graded evaluation on BloomVQA dataset with three recent VLP models~\cite{CLIP,BLIP,BLIP2} and a latest multi-modal LLM~\cite{gpt4}. We first compare the model performance on the core set of 1200 samples considering text inputs only ("QA") and multi-modal inputs ("VQA"). It is shown that all three VLP models have consistently low performance over different Bloom's levels when making prediction based on QA similarity solely. In fact, the average performance is close to the random guess which demonstrates that the proposed dataset serves as a proper evaluation set for multi-modal data. In comparison to the QA baseline, all three VLP models demonstrate improved performance when exploiting multi-modal similarity for VQA prediction for most Bloom's levels. We have similar observation for all three VLP models such that they achieve higher accuracy for multi-modal comprehension on lower-level tasks (level 1-2) in comparison to higher-level tasks (level 3-6). For comparison between three VLP models, we observe that the recent improved model (BLIP2) achieves the most gain in low-level (Level 1) tasks while similar performance is achieved by different models for higher-level tasks. This observation supports our hypothesis that VQA tasks corresponding to high-level human comprehension are not adequately addressed by current techniques, as existing datasets and models have been focusing on tasks reflecting memorization of low-level knowledge. 
\newcommand{\nSlashM}[1][0.125em]{{n\hspace{#1}\textbackslash\hspace{#1}m}}
\begin{table*}[!ht]
\centering
\small
\setlength\tabcolsep{5pt}
\begin{subtable}{0.475\textwidth}
\centering
\begin{tabular}{c|c|c|c|c|c|c}
\hline
\nSlashM & 1 & 2 & 3 & 4 & 5 & 6 \\
\hline
\shortstack{\\1} & 100.0 & 43.1 & 31.2 & 32.1 & 32.1 & 33.0\\
2 & 55.9 & 100.0 & 38.1 & 34.5 & 36.9 & 36.9\\
3 & 52.3 & 49.2 & 100.0 & 32.3 & 30.8 & 41.5\\
4 & 52.2 & 43.3 & 31.3 & 100.0 & 35.8 & 34.3\\
5 & 55.6 & 49.2 & 31.8 & 38.1 & 100.0 & 31.8\\
6 & 56.2 & 48.4 & 42.2 & 35.9 & 31.3 & 100.0\\
\hline
\end{tabular}
\caption{CLIP}\label{tab:consist_clip}
\end{subtable}
\hfill
\begin{subtable}{0.475\textwidth}
\centering
\begin{tabular}{c|c|c|c|c|c|c}
\hline
\nSlashM & 1 & 2 & 3 & 4 & 5 & 6 \\
\hline
\shortstack{\\1} & 100.0 & 47.9 & 33.9 & 35.5 & 28.9 & 34.7\\
2 & 59.8 & 100.0 & 29.9 & 35.1 & 26.8 & 31.9\\
3 & 77.4 & 54.7 & 100.0 & 49.1 & 28.3 & 30.2\\
4 & 59.7 & 47.2 & 36.1 & 100.0 & 33.3 & 37.5\\
5 & 58.3 & 43.3 & 25.0 & 40.0 & 100.0 & 33.3\\
6 & 64.6 & 47.7 & 24.6 & 41.5 & 30.8 & 100.0\\
\hline
\end{tabular}
\caption{BLIP}\label{tab:consist_blip}
\end{subtable}
\vspace{0.5em}
\begin{subtable}{0.475\textwidth}
\centering
\begin{tabular}{c|c|c|c|c|c|c}
\hline
\nSlashM & 1 & 2 & 3 & 4 & 5 & 6 \\
\hline
\shortstack{\\1} & 100.0 & 50.9 & 27.3 & 40.9 & 37.3 & 27.3\\
2 & 56.0 & 100.0 & 26.0 & 37.0 & 40.0 & 29.0\\
3 & 56.6 & 49.1 & 100.0 & 37.7 & 37.7 & 24.5\\
4 & 62.5 & 51.4 & 27.8 & 100.0 & 41.7 & 36.1\\
5 & 56.2 & 54.8 & 27.4 & 41.1 & 100.0 & 24.7\\
6 & 52.6 & 50.9 & 22.8 & 45.6 & 31.6 & 100.0\\
\hline
\end{tabular}
\caption{BLIP2}\label{tab:consist_blip2}
\end{subtable}
\hfill
\begin{subtable}{0.475\textwidth}
\centering
\begin{tabular}{c|c|c|c|c|c|c}
\hline
\nSlashM & 1 & 2 & 3 & 4 & 5 & 6 \\
\hline
\shortstack{\\1} & 100.0 & 75.9 & 81.2 & 82.7 & 78.9 & 76.7\\
2 & 68.2 & 100.0 & 78.4 & 79.1 & 79.1 & 80.4\\
3 & 68.8 & 73.9 & 100.0 & 80.9 & 80.3 & 76.4\\
4 & 70.1 & 74.5 & 80.9 & 100.0 & 78.3 & 79.0\\
5 & 67.7 & 75.5 & 81.3 & 79.4 & 100.0 & 73.6\\
6 & 66.2 & 77.3 & 77.9 & 80.5 & 74.0 & 100.0\\
\hline
\end{tabular}
\caption{GPT-4V}\label{tab:consist_gpt4}
\end{subtable}
\caption{Model consistency: accuracy (\%) at Level m when model succeeds at Level n.
}
\label{tab:consist}
\end{table*}
\paragraph{Model Consistency: Conditional Performance}
In Table~\ref{tab:consist}, we consider the 1200 core samples and report model performance on tasks of a specific Bloom's level (m) conditional on correct prediction on the task of the same story from the same annotator at a different Bloom's level (n). With this set of results, we seek to compare the model performance with human comprehension patterns. We start with the hypothesis that comprehension at a lower level is less likely to fail when comprehension at a higher level is already achieved, as many lower-level comprehension skills are readily required in solving higher-level tasks. We observe a pattern in model performance aligned with this hypothesis in some scenarios. For example, as shown in Table~\ref{tab:consist_clip}, Level 2 accuracy given success of higher-level (Level 3-6) tasks are generally higher than Level 2 accuracy given success of lower-level (Level 1) tasks. However, this observation does not hold across different Bloom's levels and different models. For example, as shown in Table~\ref{tab:consist_blip2}, Level 5 accuracy conditioned on success of higher-level (Level 6) tasks is lower than Level 5 accuracy conditional on success of lower-level tasks (Level 1-4). The misalignment between model consistency patterns and human comprehension patterns surfaces the lack of grounding in model responses.
\begin{table}
\centering
\small
\setlength\tabcolsep{4pt}
\begin{tabular}{c c c| c c| c c| c c}
\hline
&\multicolumn{2}{c|}{\shortstack{\\CLIP}}
&\multicolumn{2}{c|}{\shortstack{\\BLIP}}&\multicolumn{2}{c|}{\shortstack{\\BLIP2}}&\multicolumn{2}{c}{\shortstack{\\GPT-4V}}\\\hline
Level&\shortstack{\\aug}&\shortstack{\\AP}&\shortstack{\\aug}&\shortstack{\\AP}&\shortstack{\\aug}&\shortstack{\\AP}&\shortstack{\\aug}&\shortstack{\\AP}\\
\hline
\shortstack{\\1} &51.3 &95.1 & 55.5&97.5& 59.5&99.1&85.0 &66.0\\
2 & 38.4& 96.2 & 43.9&91.5 &44.5&93.8&93.5&74.3\\
3 & 28.8&90.0 & 28.3& 85.5&29.8 &89.4& 81.5&77.1\\
4 & 33.8 & 85.7 &32.6&86.9& 35.2& 88.7&90.0&77.4\\
5 & 32.7& 91.0& 31.1& 91.6&32.2&91.3&88.0&77.8\\
6 & 29.6&92.1&31.0&94.7 &28.5&85.3&84.5&78.0\\
\hline
\shortstack{\\Avg.} &35.8&91.7 &37.0&91.3 &38.3&91.3&87.1 &75.1\\
\hline
\end{tabular}
\caption{Accuracy on augmented data and AP consistency on data with and without augmentation (\%).}\label{tab:aug}
\end{table}
\paragraph{Model Consistency: Augmentation}
In Table~\ref{tab:aug}, we further compare the performance of different models evaluated on VQA data samples with and without augmentation. For core data samples at each Bloom's level, we augment the data by incorporating Level 1 knowledge about the same story as the context for the original question ("aug"). We further quantify the consistency of model performance between data with and without augmentation using Average Precision ("AP") scores as described in Section~\ref{sec:aug}. Overall we observe that all three VLP models have a similar consistency level averaged over 6 different Bloom's categories. The high consistency scores indicate that model performance can be perceived with confidence as low-level distractions have small effect on the model performance. Considering consistency measured with base tasks at different Bloom's levels, we observe that Level 3 and Level 4 base tasks tend to be affected at a relatively greater scale by additional context from Level 1. This observation serves an example of model-specific comprehension patterns disclosed by proposed consistency analysis.
\paragraph{Experimental Results using GPT-4V}
As shown in Table~\ref{tab:base}, GPT-4V shows strong performance across different Bloom's levels. However, the comparison between text-only experiments using GPT-4 ("QA") and multi-modal experiments using full GPT-4V ("VQA") indicates that the high accuracy of the model may stem from biases and hallucinations. The model's gain from incorporating visual inputs decreases on tasks with increasing Bloom's levels, suggesting a tendency of the model to take shortcuts especially for tasks requiring higher-level comprehension. Specifically at Level 6, the model performance degrades when visual data is introduced. In comparison to VLP models, GPT-4V handles visual inputs at story level. This may contribute to the difference in model performance. We choose picture stories designed for young children as they have simple plots suitable for demonstrating various comprehension tasks without introducing unnecessary complexity. However, with such data there can potentially exist common-sense biases especially for tasks at higher levels which require more abstract comprehension, as the stories are often based on common sense. Therefore the higher performance of GPT-4V in comparison to other baselines can also be a sign that the model is better at leveraging common sense. In this work, instead of comparing models with respect to absolute accuracy, we focus on examine whether the model prediction is grounded to the visual data and the consistency patterns of human comprehension.

In Table~\ref{tab:consist_gpt4}, we show GPT-4V performance conditional on successful prediction on tasks about the same story at a different Bloom's level. We observe that the model consistency pattern deviates from the hypothesis made based on human comprehension patterns in multiple cases. For example, Level 5 accuracy conditional on the success on higher-level (Level 6) tasks is lower than Level 5 accuracy conditional on the success on lower-level tasks (Level 1-4). Furthermore, as shown in Table~\ref{tab:aug}, in comparison to the VLP models, GPT-4V has a lower average precision score which corresponds to lower consistency on data with and without augmentation by introducing irrelevant information. These observations suggest that, although GPT-4V is demonstrating generally improved VQA accuracy in comparison to earlier models, it may still fall short in demonstrating consistent comprehension following human intuitions.
\section{Limitations}
Constrained by the resources and availability of creative commons data, we collect a relatively small-scale dataset to serve as an evaluation set for characterization of model performance based on a theoretically-grounded taxonomy. While domain gaps may exist when evaluating models pre-trained with data from different domains, the focus of this work is not to optimize VQA accuracy in the picture story domain. Instead, we focus on providing insights on model behaviors over different types of comprehension and measuring its consistency. Our proposed hierarchical data representation and novel consistency metrics are generalizable and compatible with technologies such as prompting based data generation using LLMs which opens the opportunities for future research on scaling up the dataset. 
We collect picture story data from two resources emphasizing on Indian and African backgrounds respectively. With such inputs, we encourage cross-culture understanding. The utilization of stories designed specifically for young children further injects cultural neutrality, as the contents of the stories focus on core concepts appealing to children universally and have low cultural complexity.

\section{Conclusions}
We present a novel dataset for systematic assessment of multi-modal comprehension engines. Inspired by Bloom's Taxonomy from education research, we collect multiple-choice samples based on picture story data from creative commons resources. Each sample is labeled with a cognitive skill required in solving the multiple-choice task and is associated with a specific level of comprehension from Bloom's Taxonomy. We further propose a novel hierarchical graph representation describing knowledge extracted from picture stories at different Bloom's levels. The proposed Story Graph naturally extends the concept of scene graphs and maps data into the hierarchical taxonomy. We demonstrate that automatic data augmentation can be achieved by traversing through the underlying graph. With the proposed dataset, we can not only assess given models in a graded manner but also characterize the models with respect to consistency of their performance. The proposed data structure and metrics pave the way for a wide range of interesting future works, where hierarchical graph representation can be used to guide systematic data storage, retrieval and generation.


\section*{Acknowledgements}
This work was supported in part by NSF awards \#1909696 and \#2326491. The views and conclusions contained herein are those of the authors and should not be interpreted as representing the official policies or endorsements of any sponsor.

\bibliography{main}

\appendix

\section{Appendix}
\label{sec:appendix}
\renewcommand{\thefigure}{A\arabic{figure}}
\setcounter{figure}{0}
\renewcommand{\thetable}{A\arabic{table}}
\setcounter{table}{0}
\begin{sidewaystable}[htp]
\centering
\small
\setlength\tabcolsep{5pt}
\begin{tabular}{c|c|c}
\hline
Level & \shortstack{\\Skill} & \shortstack{\\Template} \\
\hline
\shortstack{\\1} & remember & What is/are <character><action><location>? \\
\shortstack{\\2} & clarify & How would you clarify the meaning of <character><action><location>?\\
\shortstack{\\3} & apply & What would you do to change the <attribute> of <character>?\\
\shortstack{\\4} & compare & How would you compare the <attribute> of <character> before and after <character><action><location>? \\
\shortstack{\\5} & critique & How would you critique the strategy <character> used to <action>?\\
\shortstack{\\6} & improve & How would you improve the strategy <character> used to <action>?\\
\hline
\end{tabular}
\caption{Examples of templates and corresponding cognitive skills at different Bloom's levels.
}
\label{tab:template}
\end{sidewaystable}
\begin{figure*}[!h]
\includegraphics[scale=0.7]{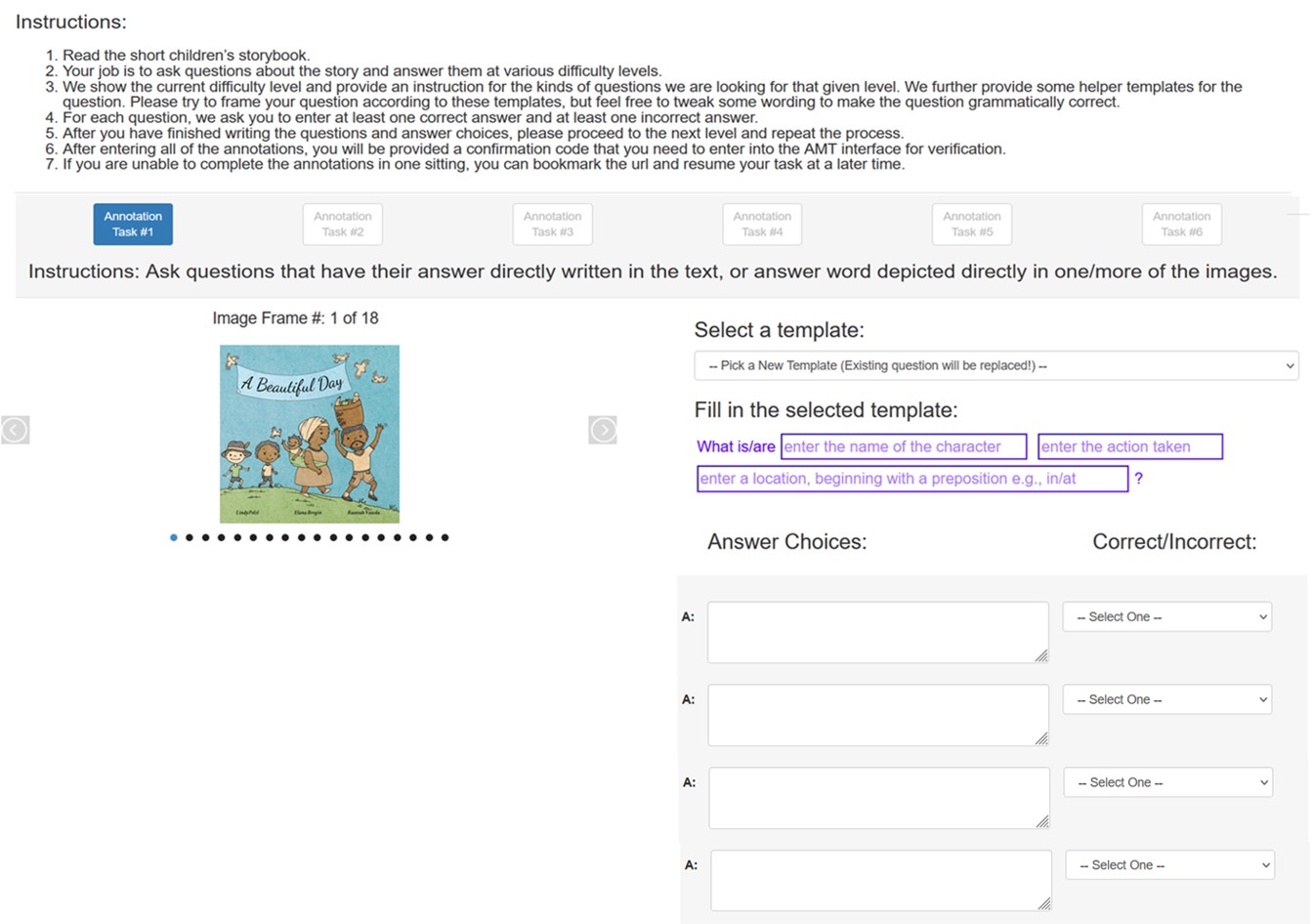}
\caption{We designed a Web-based UI with general instructions and detailed instructions at each Bloom's level provided to annotators without background expertise in the domain.} 
\label{fig:ui}
\vspace{-1.5em}
\end{figure*}
\subsection{Data Collection}
We provide a web-based UI for data collection as shown in Figure.~\ref{fig:ui}. We recruit annotators in the United States via Amazon Mechanical Turk. While our tasks are complicated, we empirically adjust our instructions, UI design and collection protocols with iterations of data collection where we increase the payment based on the quality of the inputs.
We report in Table~\ref{tab:template} the examples of templates constructed based on corresponding cognitive skills from Bloom's Taxonomy for the data collection.

\subsection{Example data}
In Figure~\ref{fig:story_example}, we show a complete picture story as one example from BloomVQA. In Table~\ref{tab:data_example}, we show a corresponding set of example questions and answers from different Bloom's levels collected based on the story. 
\begin{figure*}
\begin{tabular}{cccc}
\includegraphics[scale=0.09]{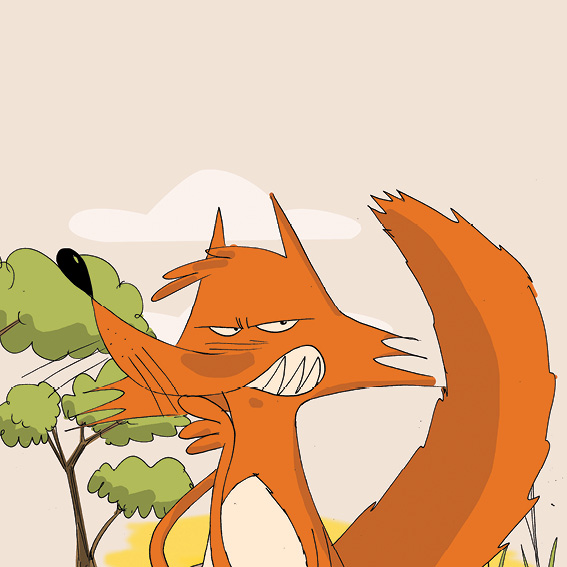} & \includegraphics[scale=0.09]{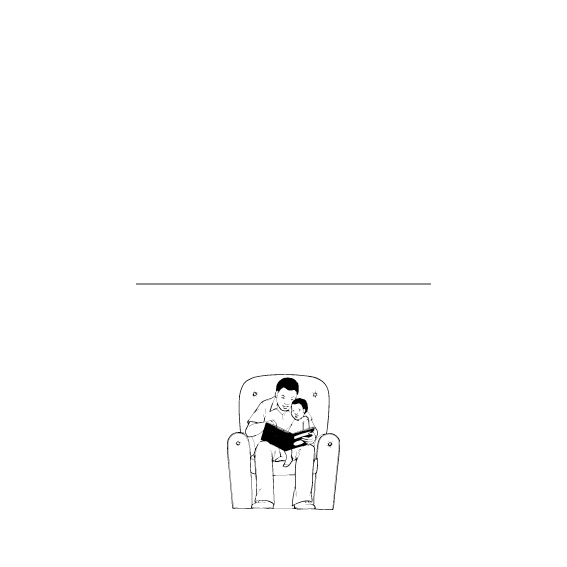} &
\includegraphics[scale=0.09]{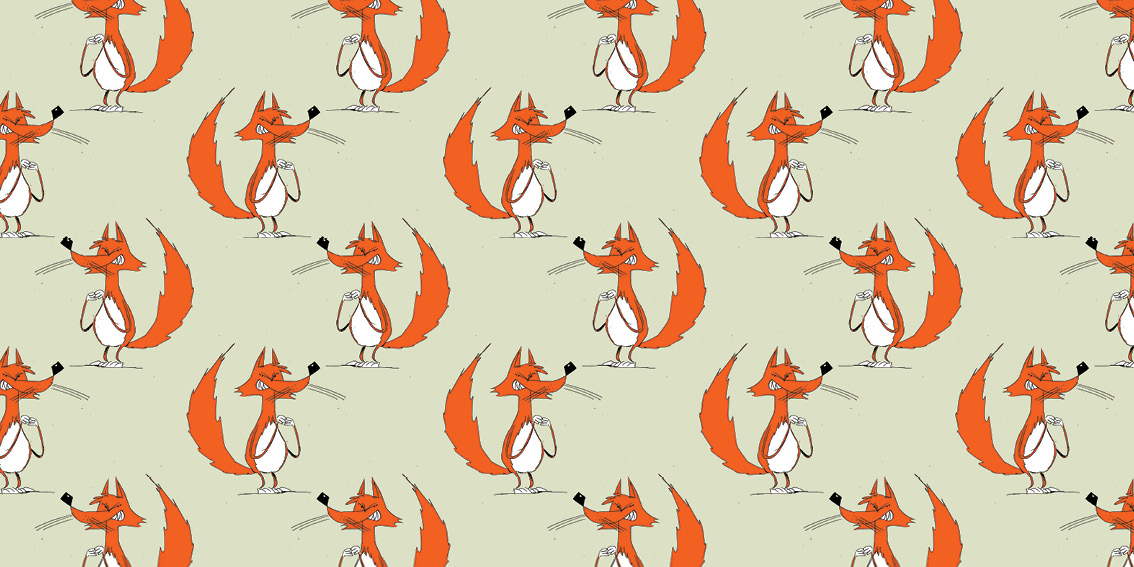} & \includegraphics[scale=0.09]{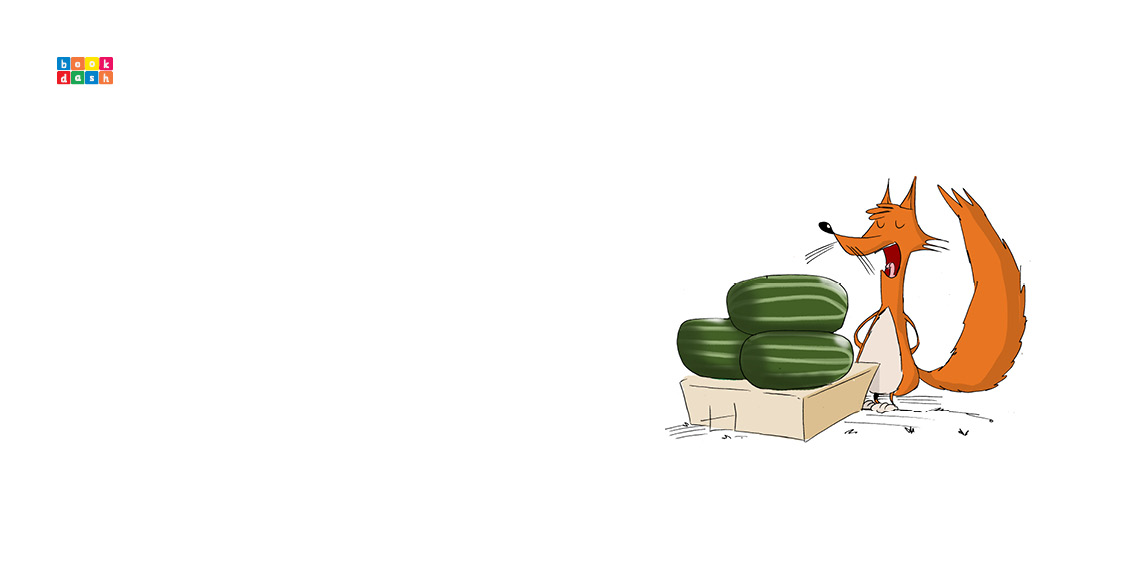} \\
\includegraphics[scale=0.09]{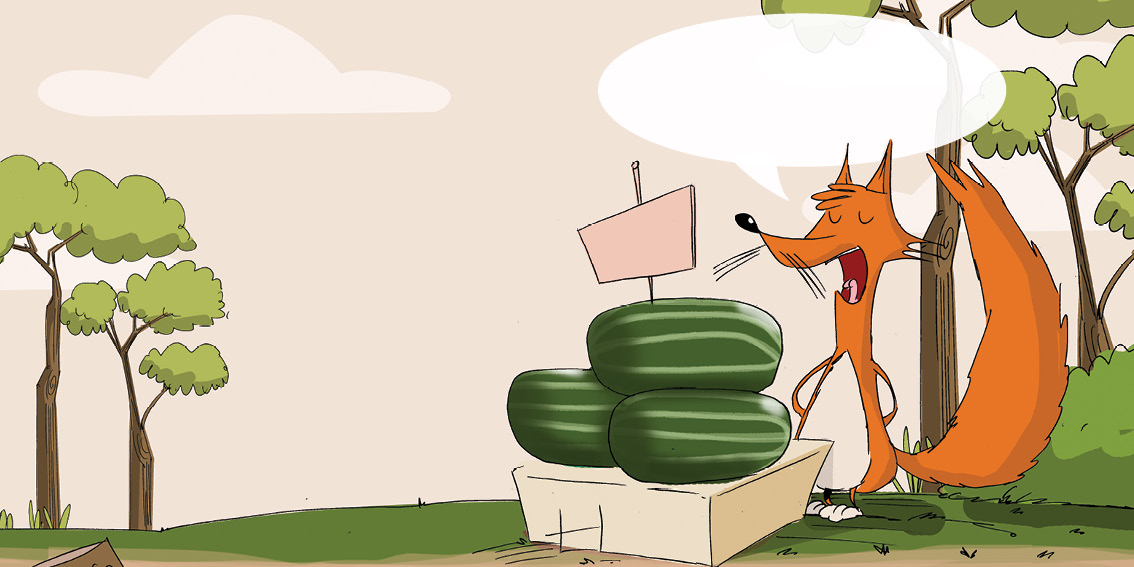} &
\includegraphics[scale=0.09]{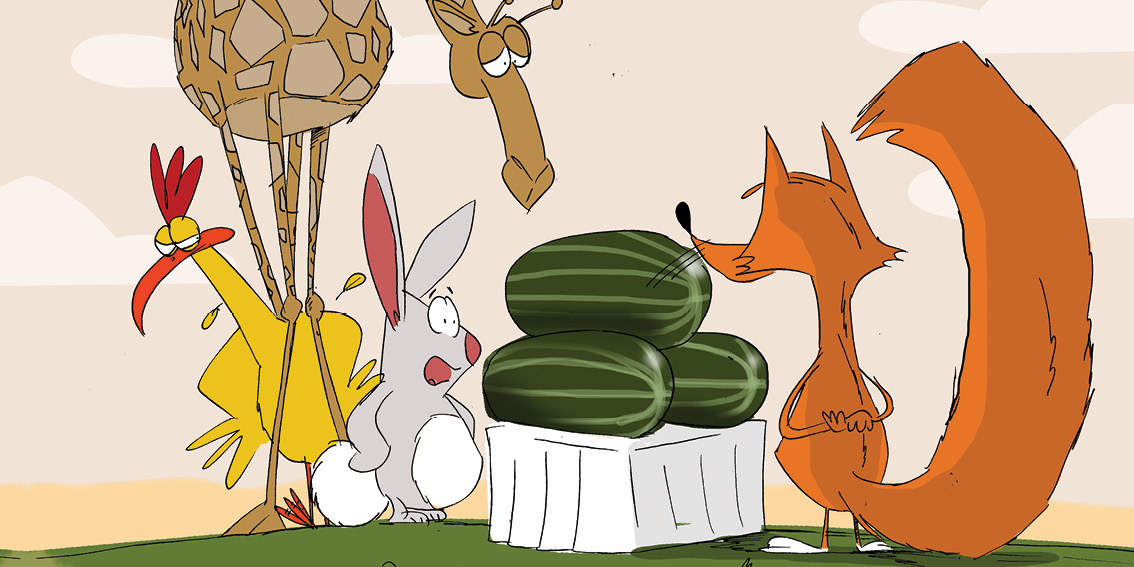} & \includegraphics[scale=0.09]{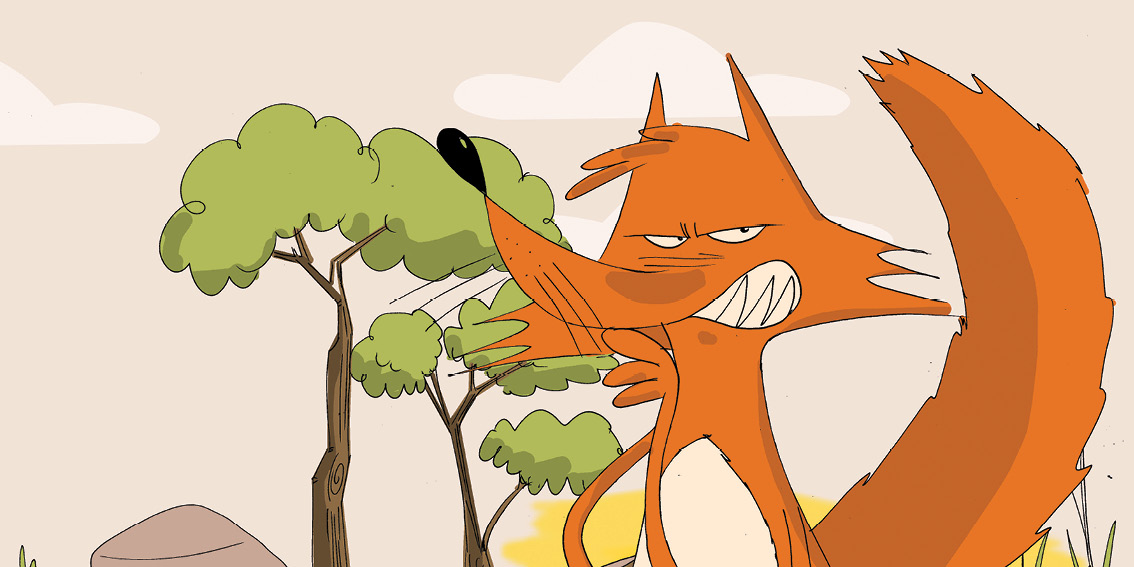} &
\includegraphics[scale=0.09]{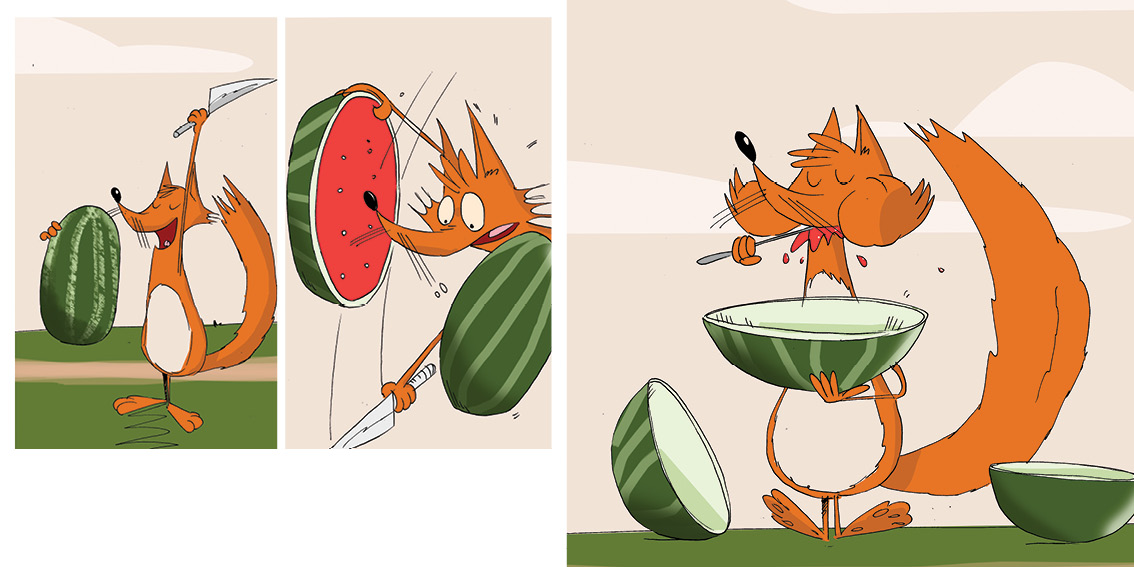} \\ \includegraphics[scale=0.09]{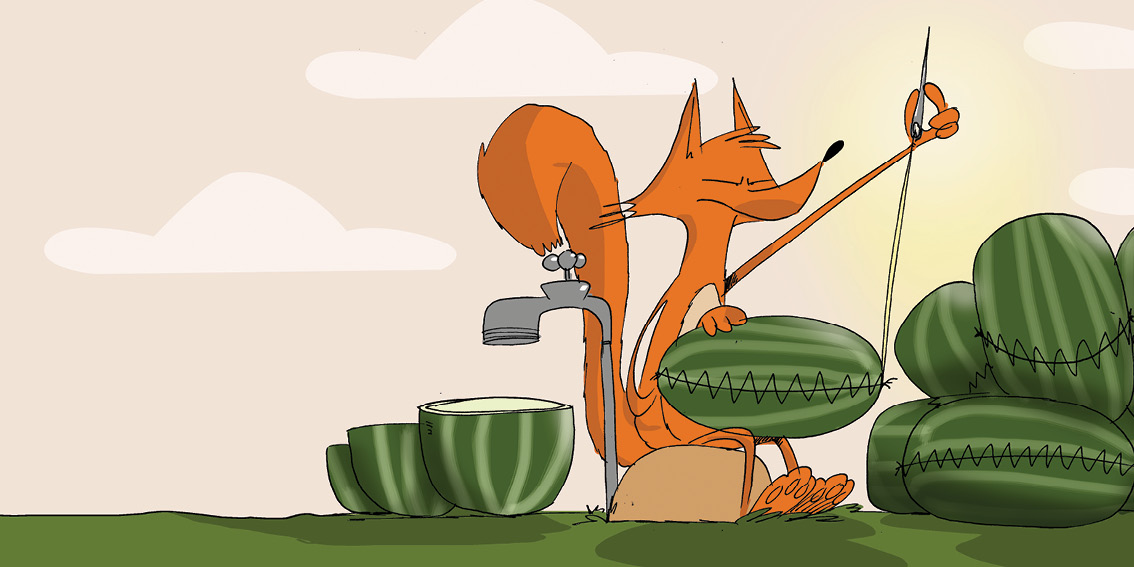} &
\includegraphics[scale=0.09]{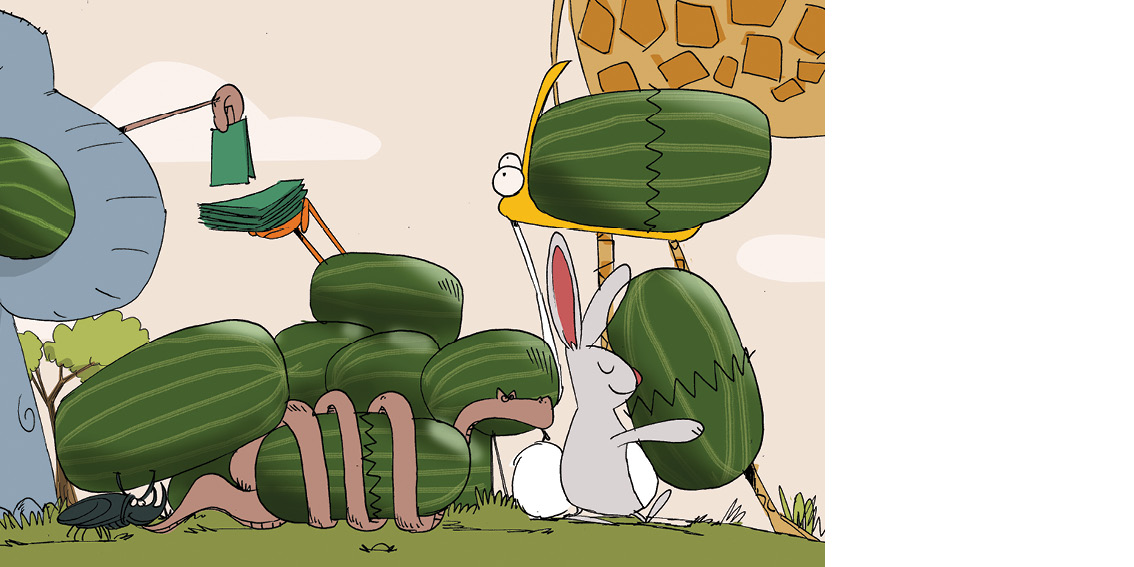} &
\includegraphics[scale=0.09]{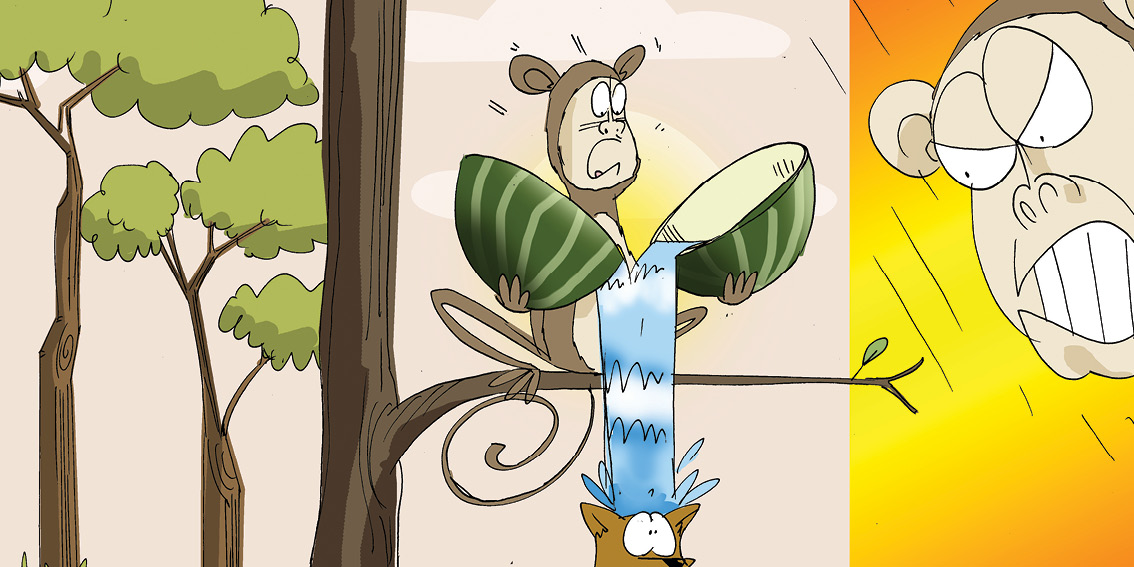} &
\includegraphics[scale=0.09]{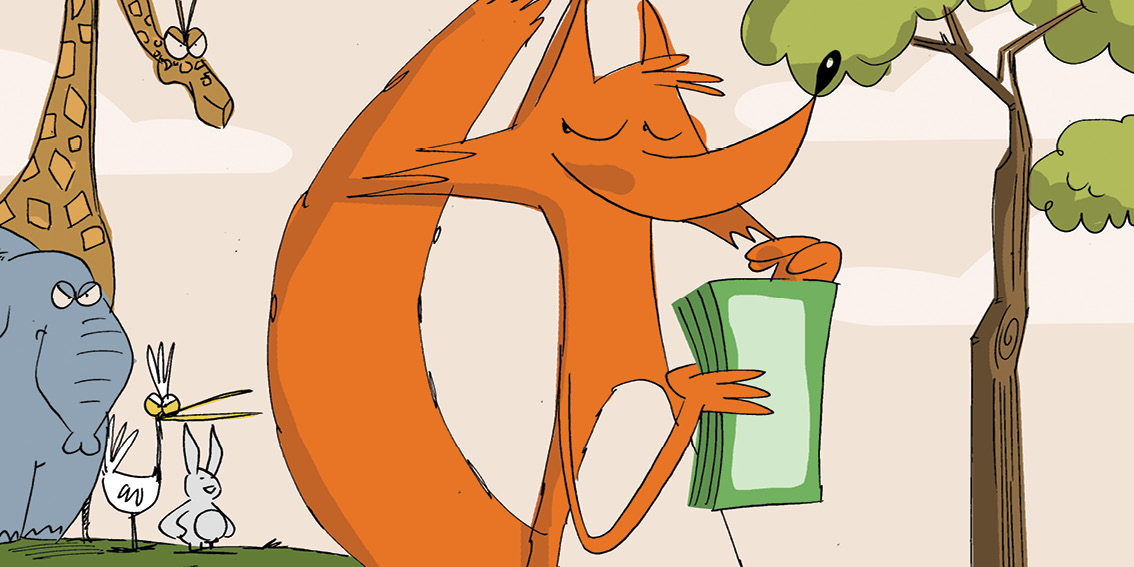} \\
\includegraphics[scale=0.09]{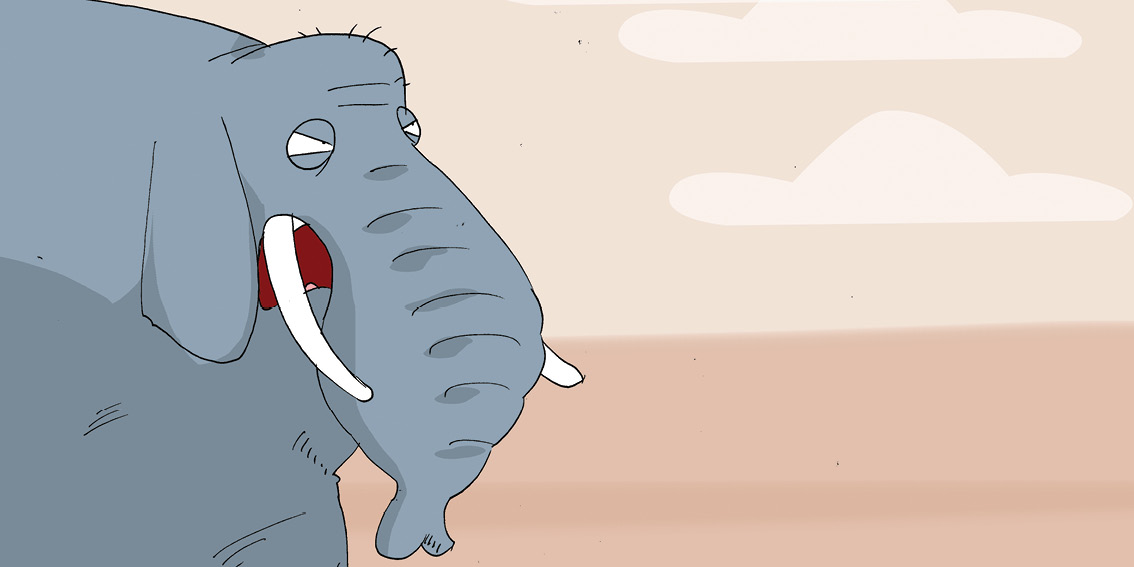} &
\includegraphics[scale=0.09]{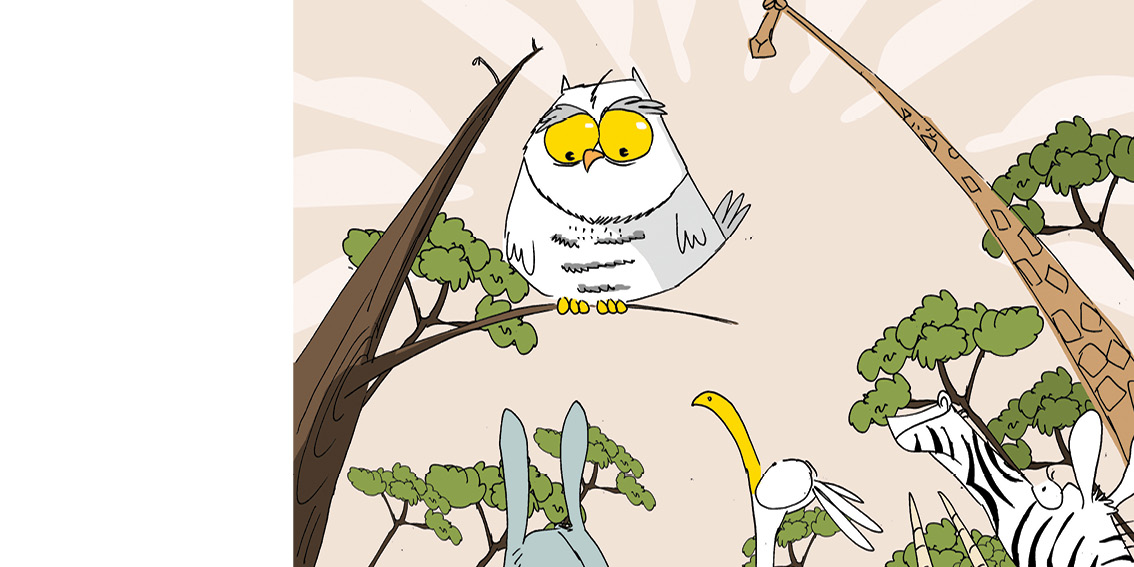} &
\includegraphics[scale=0.09]{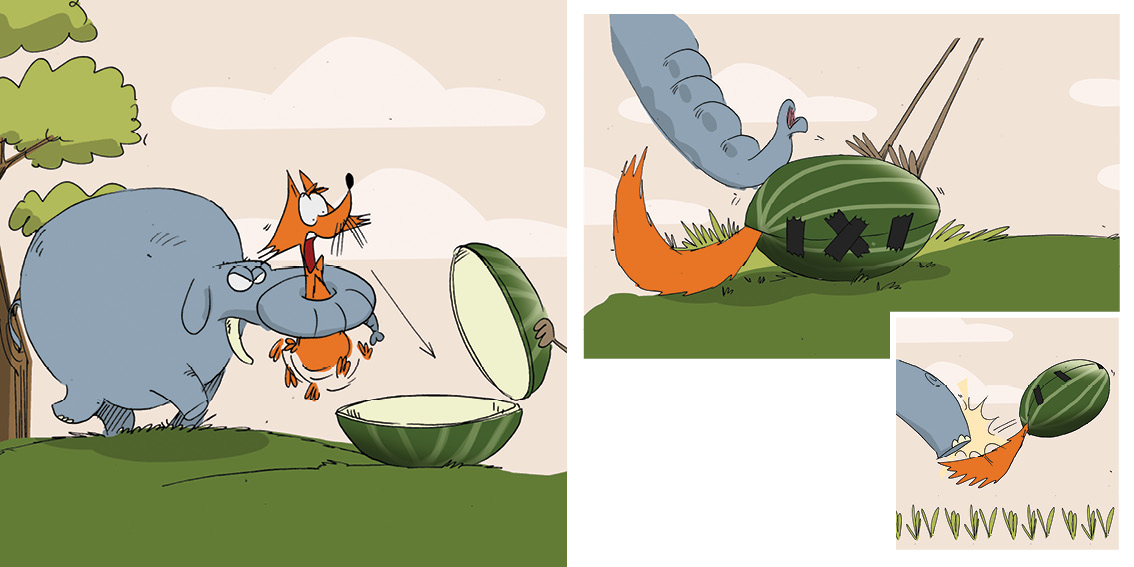} &
\includegraphics[scale=0.09]{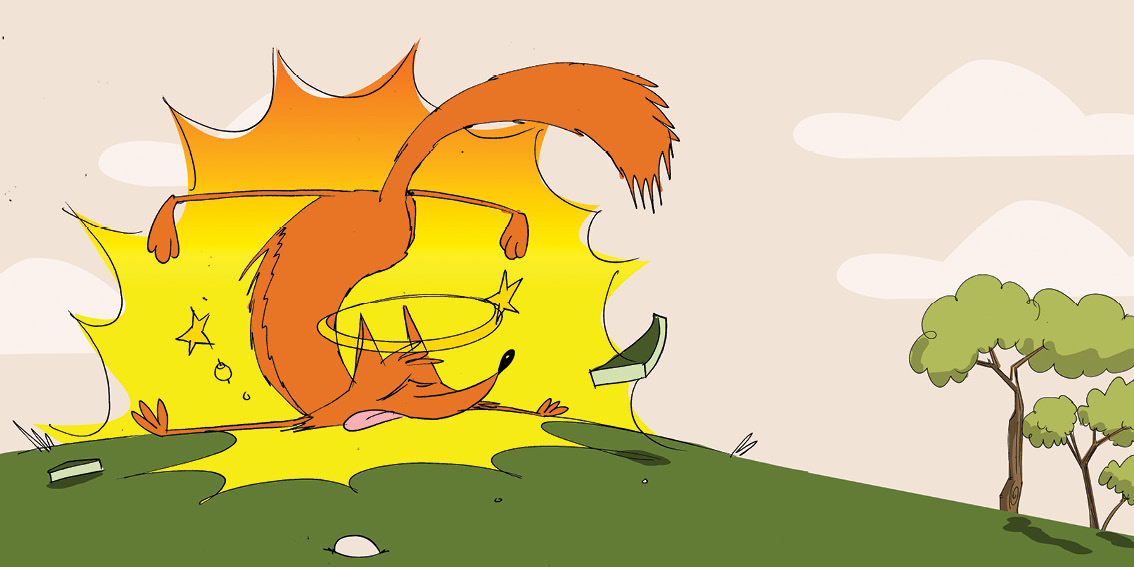} \\
\includegraphics[scale=0.09]{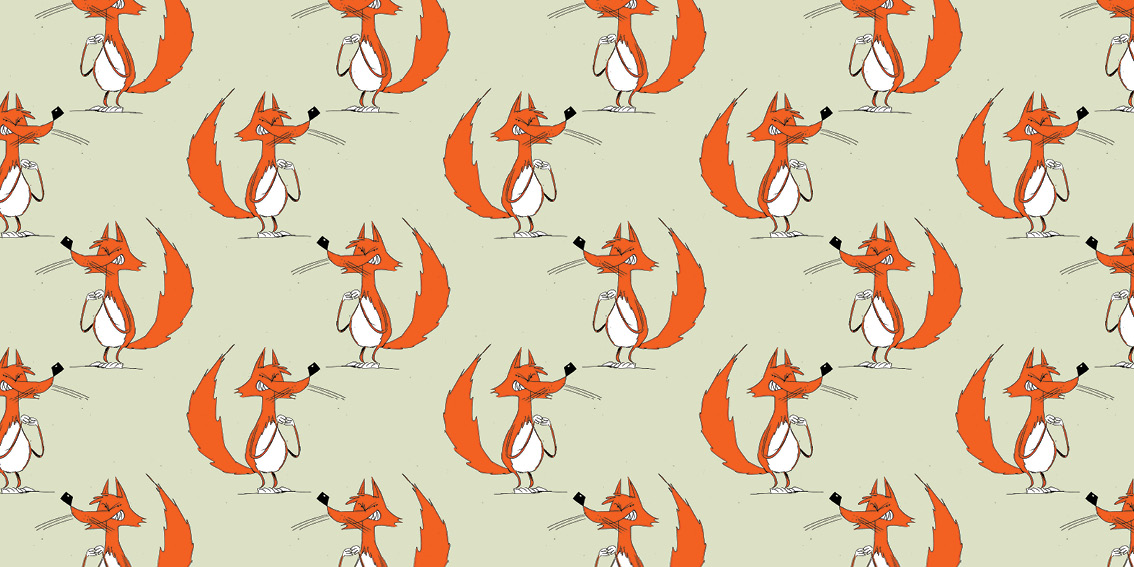} &
\includegraphics[scale=0.09]{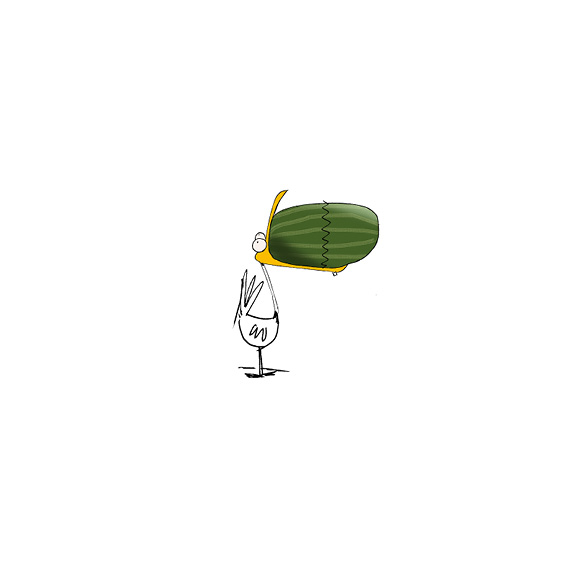} 
\end{tabular}
\caption{Example picture story: "Foxy joxy plays a trick"} 
\end{figure*}\label{fig:story_example}

\begin{sidewaystable}[htp]
\centering
\small
\vspace{20em}
\setlength\tabcolsep{5pt}
\begin{tabular}{c|c|c|c}
\hline
Level & \shortstack{\\Skill} &\shortstack{\\Question} & \shortstack{\\Answers} \\
\hline
\shortstack{\\1} & remember & What is/are Foxy Joxy selling in forest ? & \shortstack{\\fake watermelons (Correct)\\free watermelons (Incorrect)\\real watermelons (Incorrect)\\fresh fruits (Incorrect)} \\\hline
\shortstack{\\2} & clarify & Can you describe what happened when Foxy Joxy selling watermelons in the forest? & \shortstack{\\he sold at high price (Correct)\\he used no tricks (Incorrect)\\he was honest (Incorrect)\\he made the elephant happy (Incorrect)}\\\hline
\shortstack{\\3} & apply & What would you choose if Joxy tried to sell you watermelons at surprisingly low price in the forest? &\shortstack{\\doubt about the price (Correct)\\buy it without questions (Incorrect)\\buy it and then sell it to others (Incorrect)\\behave like other animals (Incorrect)}\\\hline
\shortstack{\\4} & analyze & What is the function of animal gathering? & 
\shortstack{\\coming up a plan to punish Joxy (Correct)\\gathering without the owl (Incorrect)\\discuss how to care about Joxy (Incorrect)\\take a lesson about economic (Incorrect)}\\\hline
\shortstack{\\5} & evaluate & Evaluate whether union of animal is successful or not to achieve strength of unity. & \shortstack{\\yes because they taught Joxy a lesson (Correct)\\No because Joxy never cheated (Incorrect)\\ No because they got fluffy watermelon (Incorrect)\\ Yes because the animals were well paid (Incorrect)}\\\hline
\shortstack{\\6} & improve & How would you develop a different way to solve animal's problem: got cheated by Joxy. &\shortstack{\\send him to police (Correct)\\buy more watermelons (Incorrect)\\buy other fruits (Incorrect)\\never eat watermelon again (Incorrect)}\\
\hline
\end{tabular}
\caption{Example questions and answers over different Bloom's levels on the picture story "Foxy joxy plays a trick"
}
\label{tab:data_example}
\end{sidewaystable}

\end{document}